# Parallel Spell-Checking Algorithm Based on Yahoo! N-Grams Dataset

**Youssef Bassil**

LACSC – Lebanese Association for Computational Sciences
Registered under No. 957, 2011, Beirut, Lebanon

Email: youssef.bassil@lacsc.org

**Abstract** – Spell-checking is the process of detecting and sometimes providing suggestions for incorrectly spelled words in a text. Basically, the larger the dictionary of a spell-checker is, the higher is the error detection rate; otherwise, misspellings would pass undetected. Unfortunately, traditional dictionaries suffer from out-of-vocabulary and data sparseness problems as they do not encompass large vocabulary of words indispensable to cover proper names, domain-specific terms, technical jargons, special acronyms, and terminologies. As a result, spell-checkers will incur low error detection and correction rate and will fail to flag all errors in the text. This paper proposes a new parallel shared-memory spell-checking algorithm that uses rich real-world word statistics from Yahoo! N-Grams Dataset to correct non-word and real-word errors in computer text. Essentially, the proposed algorithm can be divided into three sub-algorithms that run in a parallel fashion: The error detection algorithm that detects misspellings, the candidates generation algorithm that generates correction suggestions, and the error correction algorithm that performs contextual error correction. Experiments conducted on a set of text articles containing misspellings, showed a remarkable spelling error correction rate that resulted in a radical reduction of both non-word and real-word errors in electronic text. In a further study, the proposed algorithm is to be optimized for message-passing systems so as to become more flexible and less costly to scale over distributed machines.

## 1. Introduction

Since their inception, computers have been exploited broadly to solve and automate complex problems related to diverse domains and fields including mathematics, sciences, education, medicine, gaming, multimedia, and linguistics. In effect, computational linguistics also known as natural language processing (*NLP*) is a field of both computer science and linguistics that deals with the analysis and processing of human languages using digital computers [1]. *NLP* has also many applications, they include but not limited to Automatic Summarization, Machine Translation, Part-of-Speech Tagging (*POS*), Speech Recognition (*ASR*), Optical Character Recognition (*OCR*), and Information Retrieval (*IR*). Spell-checking is yet another significant application of computational linguistics whose research extends back to the early seventies when Ralph Gorin built the first spell-checker for the DEC PDP-10 mainframe computer at Stanford University [2]. By definition, a spell-checker is a computer program that detects and often corrects misspelled words in a text document [3]. It can be a standalone application or an add-on module integrated into an existing program such as a word processor or search engine. Fundamentally, a spell-checker is made out of three components: An error detector that detects misspelled words, a candidate spellings generator that provides spelling suggestions for the detected errors, and an error corrector that chooses the best correction out of the list of candidate spellings. All these three basic components are usually connected underneath to an internal dictionary of words that they use to validate and look-up words present in the text to be spell-checked. However, as human languages are complex and contain countless words and terms, as well as domain-specific idioms, proper names, technical terminologies, and special jargons, regular dictionaries are insufficient to cover all words in the vocabulary of the language. A problem formally known as *OOV* short for Out of Vocabulary or Data Sparseness [4] which regularly leads to false-positive and false-negative detection of out-of-dictionary words.

This paper proposes a parallel spell-checking algorithm for detecting and correcting spelling errors in computer text, based on information from Yahoo! N-Grams Dataset 2.0 [5] which embraces a substantial volume of n-gram word sequences varying between unigrams (1-gram), useful to build a lexicon model; and 5-grams useful to simulate a universal text corpus model of infinite words and expressions. Characteristically, the proposed algorithm is a shared memory model that allows for concurrent threads to execute in parallel over multi-processor or multi-core computer machines. Basically, in parallel computing, large problems can be broken down into smaller ones, and then solved simultaneously so as to achieve high-performance computing [6]. The proposed algorithm consists of three components that run in a parallel fashion: An error detector that detects non-word errors using unigrams information from Yahoo! N-Grams Dataset; a candidates generator based on a letter-based 2-gram model that generates candidates for the detected errors; and a context-sensitive error corrector that selects the best spelling candidate for correction using 5-grams information from Yahoo! N-Grams Dataset.

## 2. Spelling Correction

A spelling error is defined as $E$ in a given query word $q$ that is not an entry in a given dictionary $D$. The most basic algorithm used for spelling correction can be outlined as follows [7]:

```
SpellCorrect(word w)
{
   if (isMistake( w ))
   {
     Candidates = getCandidates( w )
     Suggestions = filterAndRank( Candidates )
     return Suggestions
   }
   else return IS CORRECT
```

First, spell correctors perform spell-checking before providing spelling suggestions. This is to avoid generating suggestions for correct words as this process is computationally intensive. Second, candidate words are generated and ranked. A candidate is a word that is the most likely to be considered as a correction for the detected error. This process results in sometimes hundreds, even thousands of candidate words. For this reason, candidates are ranked according to an internal algorithm that assigns a score or weight to every candidate. The top or highest scoring candidates are considered as real spelling suggestions.

The foremost purpose of spell-correctors is to detect and correct spelling errors which roughly range in typewritten text between 1% and 3% [8], [9], around 80% of which have one error letter, due to either transposition of two letters, adding extra letter, omitting one letter, or mistyping one letter [11]. This simple assumption makes the correction word at most one character different from its misspelled counterpart. Some experiments revealed that single-error misspellings are between 70% and 95% of all misspellings depending on the text being spell-checked [9]. Another observation concluded that 7.8% of spelling errors has the very first letter incorrect compared to 11.7% for the second letter and 19.2% for the third letter [10].

Spelling errors can be brought down into several types [9]: non-word errors which are error words that are non-words, that is, words that cannot be found in a dictionary; and real-word errors which are error words that are valid words in the dictionary but invalid with respect to their context. For instance, "aple" is a non-word error, while "from" in "fill out the from" is a real-word error. Additionally, three different types of non-word errors exist and they are [9]: mistyping, which results from manual error related to the keyboard or typewriter, e.g. "necessary" mistyped as "mecessary"; cognitive error, which results when the writer does not know how to spell the word correctly, e.g. "necessary" mistyped as "nessecary"; and phonetic error, which results from the phonetic substitution of sequence of characters with another incorrect sequence, e.g. "parachute" mistyped as "parashoote". On the other hand, another research showed that the source of spelling errors are four edit basic operations [11]: deletion when one or more characters are omitted, e.g. "tour" mistyped as "tor"; insertion when one or more characters are added, e.g. "tour" mistyped as "touur"; substitution when one or more characters are replaced with another, e.g. "tour" mistyped as "toor"; and transposition when two or more characters exchange places, e.g. "about" mistyped as "abuot".

## 3. State-of-the-Art

Work on identifying misspellings in digital text goes back to the very earliest days when text began to be manipulated by computers. Since then, spell-checking of computer text has been researched and studied thoroughly in order to improve its effectiveness and performance. Several techniques and algorithms have been conceived; the Soundex algorithm, the Bayesian model, the n-gram model, and the minimum edit distance algorithm are few to mention.

### 3.1. The Soundex Algorithm

Soundex is an algorithm for indexing words based on their phonetic sound. The cornerstone of this approach is that homophones (homophones are words with similar pronunciation but different meaning) are coded similarly so that they can be matched regardless of trivial differences in their spelling. Two strings are considered identical if they have identical Soundex Code and considered not identical otherwise. The Soundex algorithm patented in 1918 [12] converts any string of words into a code using the following rules:

- The Soundex code starts with the first letter of the string which is the only letter not to be encoded.
- The remaining letters are converted based on the following rules:

  a, e, h, i, o, u, y, w → 0
  b, f, p, v → 1
  c, g, j, k, q, s, x, z → 2
  d, t → 3
  l → 4
  m, n → 5
  r → 6

- Similar adjacent numbers are coded as a single number. e.g. change 22 to 2
- Numbers converted into '0' are removed.
- The result must be exactly one letter and three numbers. Additional letters are disregarded, while 0s are added if less than 3 numbers were obtained.

For instance, Soundex(''Robert'') = R01063 = R163 and Soundex(''Around'') = A60051 = A651

### 3.2. The Bayesian-Noisy Channel Model

The concept behind the noisy channel model is to consider a misspelling as a noisy signal which has been distorted in some way during communication. The idea behind this approach is that if one could identify how the original word was distorted, it is then straightforward to deduce the actual correction [13]. The noisy channel model is based on Bayesian inference [14] which examines some observations and classifies them into the proper categories. Studies done by Bledsoe and Browning [15], and Mosteller and Wallace [16] were the very first researches to apply the Bayesian inference to detect misspellings in electronic text.

At heart, the Bayesian model is a probabilistic model based on statistical assumptions which employs two types of probabilities: the prior probability $P(w)$ and the likelihood probability $P(O/w)$ which can be calculated as follows:

$$\hat{w} = \underset{w \in V}{\mathrm{argmax}} \frac{P(O|w)P(w)}{P(O)} = \underset{w \in V}{\mathrm{argmax}} P(O|w) P(w)$$

*P(w)* is called the prior probability and denotes the probability of *w* to occur in a particular corpus. *P(O|w)* is called the likelihood probability and denotes the probability of observing a misspelled word *O* given that the correct word is *w*. For every candidate *w*, the product of *P(O|w)*P(w)* is calculated and the one having the highest product is chosen as *w'* to correct *O*.

On the other hand, the prior probability *P(w)* is simply calculated as *P(w) = C(w) + 0.5 / N + 0.5* ; where *C(w)* is the frequency of the word *w* in the corpus, and *N* is the total number of words in the corpus. To prevent zero counts for *C(w)*, the value of 0.5 is added to the equation. In contrast, the likelihood probability *P(O|w)* is harder to compute than *P(w)* as it is normally vague to find the probability of a word to be misspelled. Nonetheless, *P(O|w)* can be estimated by calculating the probability of possible insertion, deletion, substitution, and transposition errors. Experiments carried out on the Bayesian model showed that the model can sometimes yield to incorrect results, for instance, correcting the misspelling "acress" as "acres", instead of "actress" [17].

### 3.3. The N-Gram Model

Essentially, the *n*-gram model is a probabilistic model originally devised by the Russian mathematician Andrey Markov in the early 20th century [18] and later extensively experimented by Shannon [19], Chomsky [20], and Chomsky [21] for predicting the next item in a sequence of items. The items can be letters, words, phrases, or any linguistic entity according to the application. Predominantly, the *n*-gram model is word-based used for predicting the next word in a particular sequence of words. In that sense, an *n*-gram is simply a collocation of words that is *n* words long. For instance, an *n*-gram of size 1 is referred to as a unigram; size 2 to a bigram; size 3 to a "trigram"; and so forth.

Unlike the prior probability *P(w)* which calculates the probability of a word *w* irrespective of its neighboring words, the *n*-gram model calculates the conditional probability *P(w|g)* of a word *w* given the previous sequence of words *g*. In other terms, it predicts the next word based on the foregoing *n-1* words. For instance, finding the conditional probability of *P(cat|black)* consists of calculating the probability of the entire sequence "black cat". In other words, for the word "black", the probability that the next word is "cat" is to be calculated.

Since it is too complicated to calculate the probability of a word given all previous sequence of words, the 2-gram model is used instead. It is denoted by $P(w_n|w_{n-1})$ designating the probability of a word $w_n$ given the previous word $w_{n-1}$. For a sentence enclosing a sequence of 2-gram words, the probability *P(w)* is calculated using the following equation:

$$P(w_1^n) \approx \prod_{k=1}^{n} P(w_k | w_{k-1})$$

Various investigations were conducted to improve the *n*-gram model from different aspects: Jeffreys [22], and Church and Gale [23] proposed smoothing techniques to solve the problematic of zero-frequency of *n*-grams that never occurred in a corpus; Kuhn and De Mori [24] proposed the weighted *n*-gram model which precisely approximates the *n*-grams length based on their position in the context; Niesler and Woodland [25] proposed the variable length *n*-gram model which changes the *n* size of n-grams depending on the text being manipulated so that better overall system accuracy is achieved.

### 3.4. Minimum Edit Distance

The Minimum Edit Distance algorithm [26] is defined as the minimum number of edit operations needed to transform a string of characters into another one. These operations can be identified as insertion, deletion, transpose, and substitution. In spell-checking, the goal of Minimum Edit Distance is to eliminate candidate spellings that have the largest edit distance with respect to the error word as they are regarded as sharing fewer letters with the error word than other candidates. There exist different edit distance algorithms, the most known are Levenshtein [27], Hamming [28], and Longest Common Subsequence [29].

The Levenshtein algorithm [27] named after its inventor Vladimir Levenshtein, uses a weighting approach to assign a cost of 1 to every edit operation irrespective of its type (insertion, deletion, or substitution). For instance, the Levenshtein Edit Distance between "sky" and "art" is 3 (substituting s by a, k by r, and y by t). The Levenshtein Edit Distance between "rick" and "rocky" is 2 (substituting i by o, and inserting y at the end).

The Hamming algorithm [28] is used to measure the distance between two strings of same length. It is calculated by finding the minimum number of substitutions required to transform string *x* into string *y*. For instance, the Hamming distance between "rick" and "rock" is 1 (changing i to o), and the Hamming distance between "178903" and "178206" is 2 (changing 9 to 2 and 3 to 6). The Hamming algorithm can only be applied on strings of equal length, and consequently the Hamming distance between "rick" and "rocky" is invalid because "rick" is of length 4 and "rocky" is of length 5.

Another popular technique for finding the distance between two words is the LCS short for Longest Common Subsequence [29]. The idea pivots around finding the longest common subsequence of two strings. A subsequence is a series of characters, not necessary consecutive, that appear from left to right in a string. In other terms, the longest common subsequence of two strings is the maximum length of the mutual subsequence. For example, if a=00<u>768</u>9<u>70TSGTA</u>5<u>SM</u> and b=<u>768</u>0<u>70VARDSTA</u>ABC<u>ME</u>, then LCS=<u>76870STAM</u>

### 4. Problem Statement

All the aforementioned state-of-the-art techniques are at the core based on a dictionary or lexicon of words, often implemented as a hash table [30], containing a large set of terms and words collocations, necessary for performing spell-checking. However, since languages are open, a root or stem word can give rise to thousands of valid forms of words that can hardly be represented in a regular dictionary [31]. Additionally, languages have a large vocabulary of words which includes regular words in addition to domain-specific terms, technical terminologies, special expressions, and proper names. In effect, the cause of not detecting spelling errors is an insufficient dictionary that has insufficient vocabulary coverage [9]. Moreover, more than 35% of spelling errors that pass undetected are because they were not in the dictionary [10]. A phenomenon called *OOV* short for Out of Vocabulary or Data Sparseness [4] which usually leads to false-positive and false-negative detection of out-of-dictionary words. Principally, a false-positive is a word judged to be misspelled, however, it is correct. They are usually manifested as proper nouns, domain-specific terms, and other type of words that cannot be found in a traditional dictionary. Contrariwise, a false-negative is a word judged to be correct, however, it is misspelled. They are usually real-word errors and homophones that result in valid words in the dictionary such as ''their and there'', "piece and peace", and ''to and two''. In fact, obtaining large dictionaries is the only way to improve the spell-checking error detection and correction rate. Notwithstanding, it is not enough to get a larger dictionary but also a wide-ranging and comprehensive one, encompassing proper names, domain-specific terms, and other usually out-of-dictionary words.

One interesting idea is to use a free open source word list such as ''linux.words'' with 20,000 entries or the CMU (Carnegie Mellon University) word list with 120,000 words [32], or even a more massive corpora such as the North American News Corpus which contains over 500,000 unique tokens [33]. Despite their availability, these word lists do not contain word statistics and data counts about word sequences such as *n*-grams. Furthermore, with the myriad development of electronic text, half of a million entries are still insufficient to build effective statistical language models for linguistic problems such as spell-checking.

Yahoo! N-Grams Dataset 2.0 [5] is a dataset published by Yahoo! Incorporation that houses word n-grams extracted from a corpus of 14.6 million documents all made out of 126 million unique sentences and 3.4 billion words, crawled from over 12,000 public online webpages. Additionally, Yahoo! N-Grams Dataset provides statistics such as frequency of occurrence, number, and entropy for every *n*-gram type. Since Yahoo! N-Grams Dataset contains web-scale data pulled out from the Internet, it is heavily rich in real-world data encompassing dictionary words, proper names, domain-specific terms, terminologies, acronyms, technical jargons, and special expressions that can cover most of the words and their possible sequences in the language.

## 5. Proposed Solution

This paper proposes a parallel spell-checking algorithm for detecting and correcting spelling errors in computer text, based on web-scale data from Yahoo! N-Grams Dataset 2.0 [5] which incorporates a massive volume of *n*-gram word sequences. Typically, the proposed algorithm is a shared memory model composed of concurrent execution threads, designed for multi-processor and multi-core architectures. The complete proposed solution is a blend of three sub-algorithms that run in a parallel fashion: The error detection algorithm that detects non-word errors using unigrams information from Yahoo! N-Grams Dataset; a candidates generation algorithm based on a letter-based 2-gram model that generates candidates for the detected errors; and a context-sensitive error correction algorithm that selects the best spelling candidate for correction using 5-grams statistics from Yahoo! N-Grams Dataset.

*5.1. The Parallel Error Detection Algorithm*

The proposed parallel error detection algorithm processes the original text denoted by $T=\{w_1,w_2,w_3,w_n\}$ where $w$ is a word in the original text and $n$ is the total number of words in the text, in a parallel fashion so as to detect all existing errors in $T$. The process starts by first distributing all words $w$ in $T$ over the different processors of the system. If the number of processors is less than the number of words, an equal distribution is used, that is dividing the number of words over the number of processors. The formula is given as:

*$A_k = n/p$ where $A$ represents the number of words to be assigned for a particular processor $k$, $n$ is the total number of words in the original text, and $p$ is the total number of processors in the system.*

Eventually, every processor $k$ is assigned a certain number of words $A_k >= 1$, belonging to the set $T$. Then, several threads are spawned each of which is assigned to a particular processor $k$ for execution. The task of each thread is to validate every assigned word $w_k$ (possibly multiple) against all unigrams in Yahoo! N-Grams Dataset; if $w_k$ was located, then $w_k$ is correct and thereby it needs no correction. In contrast, if $w_k$ was not located in the dataset, then $w_k$ is considered misspelled, and thus it requires correction. After the execution of all threads has been terminated, every processor $k$ adds the errors that it flagged into a shared memory location denoted by $E=\{e_1,e_2,e_3,e_m\}$ where $e$ is an error word and $m$ is the total number of errors detected in the original text $T$. Figure 1 depicts the process flow of the proposed parallel error detection algorithm.

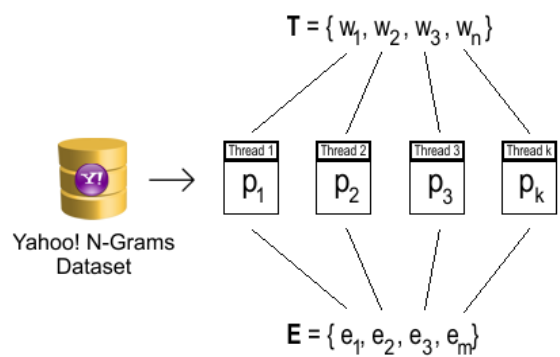

Figure 1. Process flow of the proposed parallel error detection algorithm

*5.2. The Parallel Candidates Generation Algorithm*

The proposed parallel candidates generation algorithm produces a list of possible spelling corrections or suggestions for all errors in $E$ that were detected by the error detection algorithm. Those candidates are denoted by $C=\{c_{11},c_{12},c_{13},c_{1b},...,c_{j1},c_{j2},c_{j3},c_{jd}\}$ where $c$ denotes a specific candidate spelling, $j$ denotes the candidate for the $j_{th}$ detected error, and $b$ and $d$ denote the total number of candidates generated for a particular error. Intrinsically, the algorithm exploits a letter-based 2-gram model to search in a parallel fashion for unigrams in Yahoo! N-Grams Dataset having mutual 2-gram letter sequences with the error word.

As an example, assuming that the original text to be validated is "they also work with plastic modil kits" in which the word "model" has been misspelled as "modil". Using a letter-based 2-gram model, the error word "modil" can be fragmented into 2-gram letter sequences as follows:

modil → mo , od , di , il

The task of the algorithm is to find all unigrams in Yahoo! N-Grams Dataset that contain one of the following 2-gram letter sequences: *mo, od, di,* or *il*. For this reason, every processor $k$ in the system is assigned a particular sequence $seq_k$ in $Seq=\{$"mo","od","di","il"$\}$. Then $k$ threads are spawned and executed by every processor $k$. The task of every thread $k$ is to find word unigrams in Yahoo! N-Grams Dataset that encloses $seq_k$. Assuming that the list of unigrams $L$ that was found is the following:

For $k=1$ → $seq_1$ = "mo" → $L=\{$ *mo*ld *mo*dal *mo*del *mo*m *mo*ther *mo*le $\}$
For $k=2$ → $seq_2$ = "od" → $L=\{$ m*od*al m*od*el m*od*e r*od*e tri*od*e enc*od*e $\}$
For $k=3$ → $seq_3$ = "di" → $L=\{$ la*di*ng la*di*no ra*di*an ra*di*ant *di*n para*di*ng $\}$
For $k=4$ → $seq_4$ = "il" → $L=\{$ ra*il* per*il* dera*il* ar*il* ba*il* bro*il* $\}$

The top five unigrams having the highest number of mutual 2-gram letter sequences with the error word "modil" are selected as candidates. Unigrams with equal number of common 2-gram letter sequences are ranked according to their dimension with respect to the error word, for instance, "modil" has a dimension equals to 6 (6-character-long); and thus all unigrams whose dimension is 6 are favored over those whose length is 5 or 7. Below is the list of top five unigrams based on the previous example:

$unigram_1$="modal" → 2 mutual sequences with "modil"
$unigram_2$="model" → 2 mutual sequences with "modil"
$unigram_3$="radian" → 1 mutual sequence with "modil"
$unigram_4$="mother" → 1 mutual sequence with "modil"
$unigram_5$="lading" → 1 mutual sequence with "modil"

Based on the above results, the list of generated candidates for the error word "modil" can be represented as $C_{modil}=\{$modal, model, radian, mother, lading$\}$. Next, is to select the best candidate as a correction for the error word "modil". This is in fact the actual task for the proposed parallel error correction algorithm which will be discussed in the next section. Figure 2 depicts the process flow of the proposed parallel candidates generation algorithm.

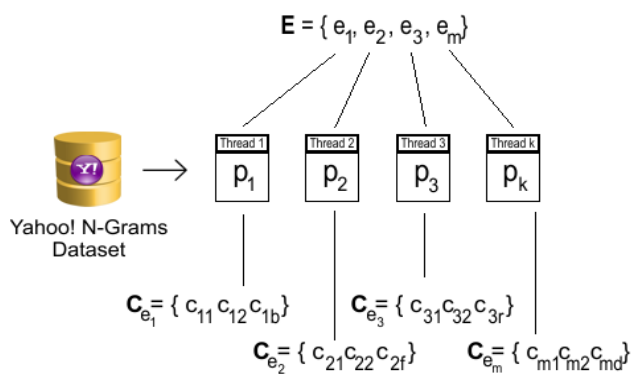

Figure 2. Process flow of the proposed parallel candidates generation algorithm

### 5.3. The Parallel Error Correction Algorithm

The proposed parallel error correction algorithm first produces several nominee sentences, each containing one of the previously generated unigram candidates with four words that initially precede the original error in $T$. Every nominee sentence can be denoted by $N_q$="$w_{q-4}$  $w_{q-3}$  $w_{q-2}$  $w_{q-1}$  $c_{qf}$" where $N$ denotes a 5-gram word nominee sentence, $w$ denotes a word preceding the original error, $c$ denotes a particular candidate spelling for the $q_{th}$ error, and $f$ denotes the $f_{th}$ candidate spelling. Then, the frequency or the number of occurrence of each created nominee sentence $N_q$ in Yahoo! N-Grams Dataset is calculated. The candidate $c_{qf}$ that belongs to the sentence $N_q$ with the highest frequency is asserted to be the real correction for the originally detected error word. The process of finding the frequency for every $N$ sentence is done in parallel. Every processor $k$ in the system is assigned a particular nominee sentence $N_q$. Then $k$ threads are spawned and executed by every processor $k$. The task of every thread $k$ is to find the frequency of $N_q$ in Yahoo! N-Grams Dataset. Back to the previous example, the list of nominee sentences $N$ can be outlined as follows:

$N_1$= "also work with plastic modal"
$N_2$= "also work with plastic model"
$N_3$= "also work with plastic radian"
$N_4$= "also work with plastic mother"
$N_5$= "also work with plastic lading"

In effect, the candidate spelling $c_q$ (either modal, model, radian, mother, or lading) in the nominee sentence $N_q$ having the highest frequency in Yahoo! N-Grams Dataset is selected as a correction for the error word "modil". The proposed algorithm is context-sensitive as it relies on real data from Yahoo! N-Grams Dataset, primarily mined from the Internet. As a result, even though the word "modal" is a valid correction for "modil", the algorithm should be able to correct it as "model" since the sentence "also work with plastic modal" is to occur very few times over the Internet, fewer than, for instance, "also work with plastic model". Figure 3 depicts the process flow of the proposed parallel error correction algorithm.

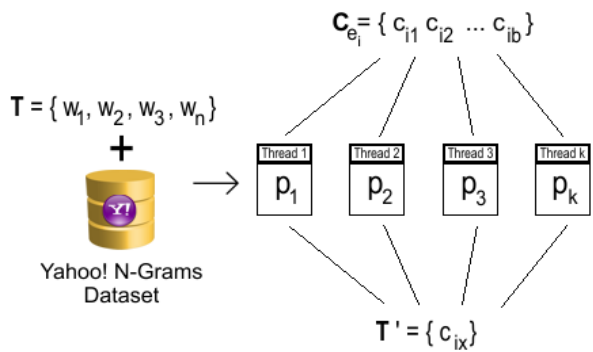

Figure 3. Process flow of the proposed parallel error correction algorithm

Below is the complete pseudo-code for the whole proposed parallel spell-checking algorithm including the error detection, candidate generation, and error correction algorithms.

```
ALGORITHM: Spell-Checking(Text)
{
    // create word-tokens out of the text to spell-check
    T ← Split(Text , " ")

    in parallel do:
      E ← spawn_threads(p , search(YahooDataSet , T[k]))
      // spawn p threads equal to the number of processors
      // search for T[k] in Yahoo data set, if found then it
      // spelled correctly;
      // otherwise it is misspelled and is stored in E

    in parallel do:
      C ← spawn_threads(p , generate_candidates(E[k]))
      // spawn p threads equal to the number of processors
      // generates candidate spelling for every error word in E.
      // store every returned candidate in C

    in parallel do:
      N ← spawn_threads(p , generate_nominees(
              T[k-4] , T[k-3] , T[k-2] , T[k-1] , C[k]))
      // spawn p threads equal to the number of processors
      // returns nominee sentences N with their frequencies in
      //  Yahoo dataset

    // returns the index of the candidate whose N has the
    // highest frequency in Yahoo dataset
    index ← max_freq(N)

    Return C[index]
    // returns the correction for the misspelled word
}
```

## 6. Experiments & Results

In the experiments, 500 articles belonging to several domains including technology, computing, economy, medicine, engineering, literature, and sports were tested. These articles encompass around 300,000 words including regular dictionary words, domain-specific terms, proper names, technical terminologies, acronyms, jargons, and expressions. Initially, those articles do not contain any misspellings; however, for evaluation purposes several words were arbitrarily changed, resulting in non-word and real-word errors in the text. These introduced misspellings were approximately 1% of the original text; and thus, they are around 3,000 spelling errors. Table 1 gives the total number of words in these selected articles, in addition to the number of introduced non-word and real-word errors.

Table 1. Number of Introduced Errors

| Total Words | 300,000 |
|---|---|
| Total Errors | 1% of 300,000=3,000 |
| Non-Word Errors | 87% of 3,000=2,600 |
| Real-Word Errors | 13% of 3,000=400 |

Spell-checking the test data using the proposed algorithm resulted in 2,831 out of 3,000 errors being corrected successfully, among which 2,571 were non-word errors and 260 were real-word errors. As a result, around 94% of total errors were corrected successfully. This includes around 99% of total non-word errors and around 65% of total real-word errors. Table 2 delineates the obtained results using the proposed algorithm.

Table 2. Test Results using the Proposed Algorithm

| Total Errors=3,000 1% of 300,000 total words | | Non-Word Errors=2,600 87% of 3,000 | | Real-Word Errors=400 13% of 3,000 | |
|---|---|---|---|---|---|
| Corrected | Not Corrected | Corrected | Not Corrected | Corrected | Not Corrected |
| 2,831 | 169 | 2,571 | 29 | 260 | 140 |
| 94% of 3,000 | 6% of 3,000 | 99% of 2,600 | 1% of 2,600 | 65% of 400 | 35% of 400 |

Below are examples of successful and unsuccessful corrections observed during the execution of the proposed error correction algorithm. It is worth noting that errors are marked by an underline and results are interpreted using a special notation in the form of [ *error-type ; corrected error ; intended word* ].

Errors Successfully Corrected 94%:

… would like to ask you to voice your sopport for this bill … → [non-word error ; support ; support]
… but the content of a computer is vulnerable to fee risks … → [real-word error ; few ; few]
… medical errors effect us all whether we are involved or not … → [real-word error ; affect ; affect]
… many of the best poems are found in too collections … → [real-word error ; two ; two]

Errors Not Corrected 2%:

… whether you hit the road in a sleek imported sporting car … → [real-word error ; sporting ; sports]
… we fear the precaution of medication prior to tonsillectomy … → [real-word error ; fear ; feel]

Errors Falsely Corrected 4%:

… After all I slept near my door on the pavement… → [real-word error ; dog ; door]
… I saw the ball running too fast … → [real-word error ; bus ; ball]

For comparison purposes, the same articles were spell-checked using two well-known spell-checkers: the free Hunspell spell-checker [34] which is the primary spell-checking tool for several of Mozilla products and OpenOffice online suite, and Ginger [35] which is a proprietary context-sensitive grammar and spell-checker. Figure 4 shows a histogram representation for the obtained results including the number of corrected spelling errors and the error correction rate.

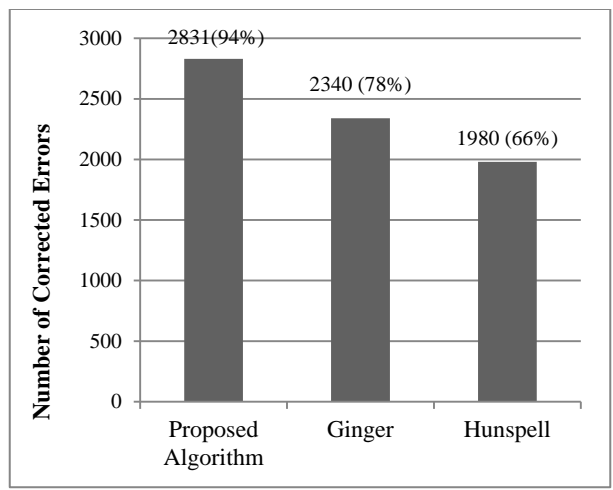

Figure 4. Number of corrected errors

Consequently, the improvement for the proposed algorithm over Ginger can be calculated as I = 2831/2340 = 1.2 = 120%, that is increasing the rate of error detection and correction by a factor of 1.2, corresponding to 20% more errors being corrected by the proposed algorithm. Likewise, the improvement for the proposed algorithm over Hunspell can be calculated as I = 2831/1980 = 1.42 = 142%, that is increasing the rate of error detection and correction by a factor of 1.42, corresponding to 42% more errors being corrected by the proposed algorithm.

## 7. Conclusions and Future Work

This paper presented a novel shared-memory parallel spell-checking algorithm for detecting and correcting spelling errors in computer text. The proposed algorithm is based on Yahoo! N-Grams Dataset that comprises trillions of word sequences and *n*-grams, originally extracted from the World Wide Web. When experimented to correct misspellings in 300,000-word articles, the proposed algorithm outclassed other existing spell-checkers as it effectively corrected 94% of the total errors, distributed as 99% non-word errors and 65% real-word errors. On the other hand, the Hunspell spell-checker managed to correct 66% of total errors; while, the Ginger spell-checker was able of 78% of total errors. In sum, the error correction rate for the proposed algorithm was 16% higher than Ginger and 28% higher than Hunspell. The major reason behind these outstanding results is the use of Yahoo! N-Grams Dataset as a dictionary model which delivers wide-ranging set of words and *n*-gram statistics that cover domain-specific terms, technical jargons, proper names, special acronyms, and most of the words that possibly can occur in a text.

As future work, a distributed message-passing algorithm is to be developed and experimented; it can typically be deployed over *n*-tier distributed computing infrastructures made out of remote servers and machines dispersed in different locations around the world. Since message-passing architectures allow computing power to be added in small increments at lower costs and at higher flexibility, it can easily be scaled for spell-checking very large electronic text.

## Acknowledgment

This research was funded by the Lebanese Association for Computational Sciences (LACSC), Beirut, Lebanon under the "Web-Scale Spell-Checking Research Project – WSSCRP2011".